\documentclass[letterpaper]{article}
\usepackage{proceed}
\usepackage[margin=1in]{geometry}
\usepackage{times}
\newtheorem{theorem}{Theorem}

\usepackage{amsmath, amsfonts}
\usepackage{graphicx}
\usepackage{xspace}

\usepackage{color}
\usepackage{algorithm}
\usepackage[noend]{algpseudocode}
\usepackage{natbib}
\usepackage{enumitem}

\title{Bayesian Inference of Log Determinants}

\author{Jack Fitzsimons$^1$ \And Kurt Cutajar$^2$ \And  Michael Osborne$^1$\\ \\
$^1$ Information Engineering, University of Oxford, UK\\
$^2$ Department of Data Science, EURECOM, France\\ 
\And  Stephen Roberts$^1$ \And Maurizio Filippone$^2$ 
}        

\begin{document}

\maketitle

\definecolor{mygreen}{rgb}{0.2, 0.7, 0.2}
\definecolor{myorange}{rgb}{0.9, 0.5, 0.0}

\newcommand\noteMF[1]{\textcolor{red}{MF - #1}}
\newcommand\noteKC[1]{\textcolor{blue}{KC - #1}}
\newcommand\noteJF[1]{\textcolor{myorange}{JF - #1}}
\newcommand\noteMO[1]{\textcolor{mygreen}{MO - #1}}

\newcommand{\nobs}{n} 
\newcommand{\R}{\mathbb{R}}
\newcommand{\N}{\mathbb{N}}
\newcommand{\Z}{\mathbb{Z}}
\newcommand{\F}{\mathcal{F}}
\newcommand{\I}{\mathcal{I}}
\newcommand{\LL}{\mathcal{L}}
\newcommand{\uu}{\mathbf{u}}
\newcommand{\ee}{\mathbf{e}}

\newcommand{\E}{\mathrm{E}}
\newcommand{\const}{\mathrm{const.}}
\newcommand{\diag}{\mathrm{diag}}
\newcommand{\Tr}{\mathrm{Tr}}
\newcommand{\Det}{\mathrm{Det}}
\newcommand{\GP}{\mathrm{GP}}
\newcommand{\ud}{\mathrm{d}}

\newcommand{\norm}{\mathcal{N}}

\newcommand{\avect}{\mathbf{a}}
\newcommand{\dvect}{\mathbf{d}}
\newcommand{\fvect}{\mathbf{f}}
\newcommand{\gvect}{\mathbf{g}}
\newcommand{\hvect}{\mathbf{h}}
\newcommand{\mvect}{\mathbf{m}}
\newcommand{\pvect}{\mathbf{p}}
\newcommand{\svect}{\mathbf{s}}
\newcommand{\uvect}{\mathbf{u}}
\newcommand{\vvect}{\mathbf{v}}
\newcommand{\zvect}{\mathbf{z}}
\newcommand{\xvect}{\mathbf{x}}
\newcommand{\yvect}{\mathbf{y}}
\newcommand{\wvect}{\mathbf{w}}
\newcommand{\Wvect}{\mathbf{W}}
\newcommand{\tvect}{\mathbf{t}}
\newcommand{\zerovect}{\mathbf{0}}
\newcommand{\onesvect}{\mathbf{1}}

\newcommand{\betavect}{\boldsymbol{\beta}}
\newcommand{\thetavect}{\boldsymbol{\theta}}
\newcommand{\Thetavect}{\mathbf{\Theta}}
\newcommand{\psivect}{\boldsymbol{\psi}}
\newcommand{\Psivect}{\boldsymbol{\Psi}}
\newcommand{\etavect}{\boldsymbol{\eta}}
\newcommand{\rhovect}{\boldsymbol{\rho}}
\newcommand{\tauvect}{\boldsymbol{\tau}}
\newcommand{\nuvect}{\boldsymbol{\nu}}
\newcommand{\muvect}{\boldsymbol{\mu}}
\newcommand{\omegavect}{\boldsymbol{\omega}}
\newcommand{\Omegavect}{\mathbf{\Omega}}
\newcommand{\sigmavect}{\boldsymbol{\sigma}}
\newcommand{\zetavect}{\boldsymbol{\zeta}}
\newcommand{\varepsilonvect}{\boldsymbol{\epsilon}}
\newcommand{\deltavect}{\boldsymbol{\delta}}

\newcommand{\bigO}{\mathcal{O}}

\newcommand{\name}[1]{{\textsc{#1}}\xspace}

\newcommand{\mcmc}{\name{mcmc}}

\newcommand{\gp}{\name{gp}}
\newcommand{\ard}{\name{ard}}

\newcommand{\relu}{{\textsc{r}}e\name{lu}}

\newcommand{\arc}{\name{arc}}
\newcommand{\rbf}{\name{rbf}}

\newcommand{\mvp}{\name{mvp}}
\newcommand{\mnll}{\name{mnll}}
\newcommand{\rmse}{\name{rmse}}
\newcommand{\nelbo}{\name{nelbo}}

\begin{abstract}
The log-determinant of a kernel matrix appears in a variety of machine learning problems, ranging from determinantal point processes and generalized Markov random fields, through to the training of Gaussian processes.
Exact calculation of this term is often intractable when the size of the kernel matrix exceeds a few thousands.
In the spirit of probabilistic numerics, we reinterpret the problem of computing the log-determinant as a Bayesian inference problem.
In particular, we combine prior knowledge in the form of bounds from matrix theory and evidence derived from stochastic trace estimation to obtain probabilistic estimates for the log-determinant and its associated uncertainty within a given computational budget.
Beyond its novelty and theoretic appeal, the performance of our proposal is competitive with state-of-the-art approaches to approximating the log-determinant, while also quantifying the uncertainty due to budget-constrained evidence.
\end{abstract}


\section{INTRODUCTION}

Developing scalable learning models without compromising performance is at the forefront of machine learning research.
The scalability of several learning models is predominantly hindered by linear algebraic operations having large computational complexity, among which is the computation of the log-determinant of a matrix~\citep{Golub96}.
The latter term features heavily in the machine learning literature, with applications including spatial models~\citep{Aune14,Rue05}, kernel-based models~\citep{Davis07,Rasmussen06}, and Bayesian learning~\citep{MacKay03}.

The standard approach for evaluating the log-determinant of a positive definite matrix involves the use of Cholesky decomposition \citep{Golub96}, which is employed in various applications of statistical models such as kernel machines. 
However, the use of Cholesky decomposition for general dense matrices requires $\mathcal{O}(n^3)$ operations, whilst also entailing memory requirements of $\mathcal{O}(n^2)$. 
In view of this computational bottleneck, various models requiring the log-determinant for inference bypass the need to compute it altogether \citep{Anitescu12,Stein13,Cutajar16,FilipponeICML15}.


Alternatively, several methods exploit sparsity and structure within the matrix itself to accelerate computations.
For example, sparsity in Gaussian Markov Random fields (GMRFs) arises from encoding conditional independence assumptions that are readily available when considering low-dimensional problems.
For such matrices, the Cholesky decompositions can be computed in fewer than $\mathcal{O}(n^3)$ operations~\citep{Rue05,Rue09}.
Similarly, Kronecker-based linear algebra techniques may be employed for kernel matrices computed on regularly spaced inputs~\citep{Saatci11}.
While these ideas have proven successful for a variety of specific applications, they cannot be extended to the case of general dense matrices without assuming special forms or structures for the available data.

To this end, general approximations to the log-determinant frequently build upon stochastic trace estimation techniques using iterative methods~\citep{Avron11}.
Two of the most widely-used polynomial approximations for large-scale matrices are the Taylor and Chebyshev expansions~\citep{Aune14,Han15}.
A more recent approach draws from the possibility of estimating the trace of functions using stochastic Lanczos quadrature~\citep{Ubaru16}, which has been shown to outperform polynomial approximations from both a theoretic and empirical perspective.

Inspired by recent developments in the field of probabilistic numerics~\citep{Hennig15}, in this work we propose an alternative approach for calculating the log-determinant of a matrix by expressing this computation as a Bayesian quadrature problem.
In doing so, we reformulate the problem of {\em computing} an intractable quantity into an {\em estimation} problem, where the goal is to infer the correct result using tractable computations that can be carried out within a given time budget. 
In particular, we model the eigenvalues of a matrix $A$ from noisy observations of $\Tr(A^k)$ obtained through stochastic trace estimation using the Taylor approximation method~\citep{Zhang07}.
Such a model can then be used to make predictions on the infinite series of the Taylor expansion, yielding the estimated value of the log-determinant.
Aside from permitting a probabilistic approach for predicting the log-determinant, this approach inherently yields uncertainty estimates for the predicted value, which in turn serves as an indicator of the quality of our approximation.

Our contributions are as follows.
\begin{enumerate}[topsep=0pt,itemsep=0.5ex]
\item We propose a probabilistic approach for computing the log-determinant of a matrix which blends different elements from the literature on estimating log-determinants under a Bayesian framework.
\item We demonstrate how bounds on the expected value of the log-determinant improve our estimates by constraining the probability distribution to lie between designated lower and upper bounds.
\item Through rigorous numerical experiments on synthetic and real data, we demonstrate how our method can yield superior approximations to competing approaches, while also having the additional benefit of uncertainty quantification.
\item Finally, in order to demonstrate how this technique may be useful within a practical scenario, we employ our method to carry out parameter selection for a large-scale determinantal point process.
\end{enumerate}

To the best of our knowledge, this is the first time that the approximation of log-determinants is viewed as a Bayesian inference problem, with the resulting quantification of uncertainty being hitherto unexplored thus far.

\subsection{RELATED WORK}

The most widely-used approaches for estimating log-determinants involve extensions of iterative algorithms, such as the Conjugate-Gradient and Lanczos methods, to obtain estimates of functions of matrices \citep{Chen11,Han15} or their trace \citep{Ubaru16}.
The idea is to rewrite log-determinants as the trace of the logarithm of the matrix, and employ trace estimation techniques~\citep{Hutchinson90} to obtain unbiased estimates of these. 
\cite{Chen11} propose an iterative algorithm to efficiently compute the product of the logarithm of a matrix with a vector, which is achieved by computing a spline approximation to the logarithm function.
A similar idea using Chebyshev polynomials has been developed by \cite{Han15}.
Most recently, the Lanczos method has been extended to handle stochastic estimates of the trace and obtain probabilistic error bounds for the approximation~\citep{Ubaru16}.
Blocking techniques, such as in \cite{Ipsen11} and \cite{Ambikasaran16}, have also been proposed.

In our work, we similarly strive to use a small number of matrix-vector products for approximating log-determinants.
However, we show that by taking a Bayesian approach we can combine priors with the evidence gathered from the intermediate results of matrix-vector products involved in the afore-mentioned methods to obtain more accurate results. 
Most importantly, our proposal has the considerable advantage that it provides a full distribution on the approximated value.


Our proposal allows for the inclusion of explicit bounds on log-determinants to constrain the posterior distribution over the estimated log-determinant~\citep{Bai97}.
Nystr\"om approximations can also be used to bound the log-determinant, as shown by \cite{Bardenet15}.
Similarly, Gaussian processes~\citep{Rasmussen06} have been formulated directly using the eigendecomposition of its spectrum, where eigenvectors are approximated using the Nystr\"om method \citep{Peng15b}.
There has also been work on estimating the distribution of kernel eigenvalues by analyzing the spectrum of linear operators \citep{Braun06,Wathen15}, as well as bounds on the spectra of matrices with particular emphasis on deriving the largest eigenvalue~\citep{Wolkowicz80,Braun06}.
In this work, we directly consider bounds on the log-determinants of matrices~\citep{Bai97}.



\section{BACKGROUND}

As highlighted in the introduction, several approaches for approximating the log-determinant of a matrix rely on stochastic trace estimation for accelerating computations.
This comes about as a result of the relationship between the log-determinant of a matrix, and the corresponding trace of the log-matrix, whereby
\begin{equation}\label{eqn:logdet_tr_identity}
\log\bigl(\Det\left(A\right)\bigr) = \Tr\bigl(\log\left(A\right)\bigr)\text{.}
\end{equation}
\noindent Provided the matrix $\log(A)$ can be efficiently sampled, this simple identity enables the use of stochastic trace estimation techniques~\citep{Avron11,Fitzsimons16}.
We elaborate further on this concept below.

\subsection{STOCHASTIC TRACE ESTIMATION}

The standard approach for computing the trace term of a matrix $A\in\mathbb{R}^{n \times n}$ involves summing the eigenvalues of the matrix.
Obtaining the eigenvalues typically involves computational complexity of $\bigO(n^3)$, which is infeasible for large matrices.
However, it is possible to obtain a stochastic estimate of the trace term such that the expectation of the estimate matches the term being approximated~\citep{Avron11}.
In this work, we shall consider the Gaussian estimator, whereby we introduce $N_\mathbf{r}$ vectors $\mathbf{r}^{(i)}$ sampled from an independently and identically distributed zero-mean and unit variance Gaussian distribution.
This yields the unbiased estimate
\begin{equation}
\Tr(A) = \frac{1}{N_r}\sum_{i=1}^{N_r}{\mathbf{r}^{(i)}}^\top A\,\mathbf{r}^{(i)}\text{.}
\end{equation}
Note that more sophisticated trace estimators~\citep[see ][]{Fitzsimons16} may also be considered; without loss of generality, we opt for a more straightforward approach in order to preserve clarity.

\subsection{TAYLOR APPROXIMATION}\label{sec:taylor}

\begin{figure}[]
\centering
	\includegraphics[width=0.5\textwidth]{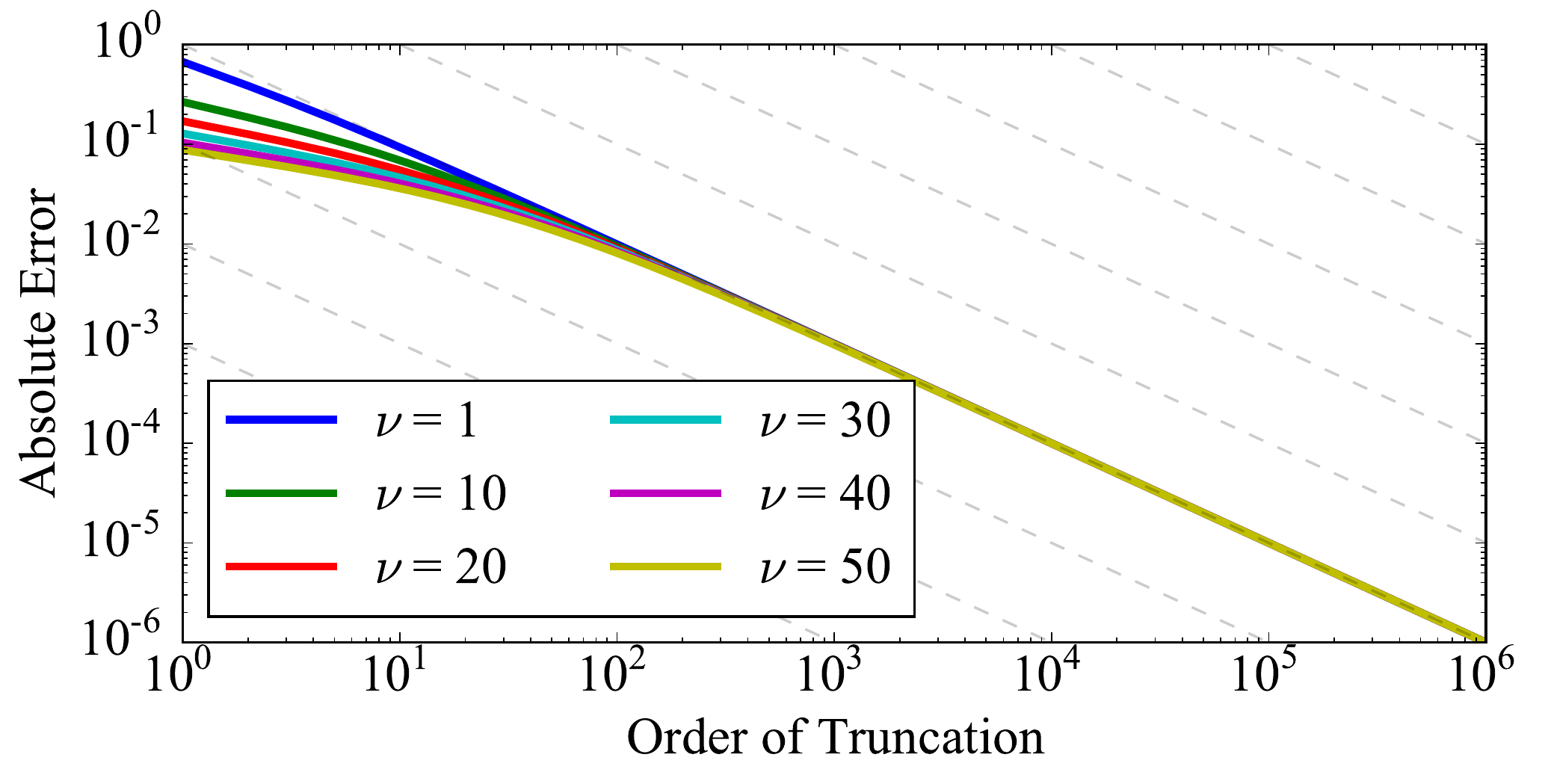}
	\caption{Expected absolute error of truncated Taylor series for stationary $\nu$-continuous kernel matrices. The dashed grey lines indicate $\mathcal{O}(n^{-1})$.}
	\label{TaylorError}
\end{figure}

Against the backdrop of machine learning applications, in this work we predominantly consider covariance matrices taking the form of a Gram matrix $K = \left\{ \kappa(\xvect_i, \xvect_j) \right\}_{i,j=1,...,n}$, where the kernel function $\kappa$ implicitly induces a feature space representation of data points $\xvect_i$.
Assume $K$ has been normalized such that the maximum eigenvalue is less than or equal to one, $\lambda_{0}\leq1$, where the largest eigenvalue can be efficiently found using Gershgorin intervals~\citep[][]{Gershgorin31}.
Given that covariance matrices are positive semidefinite, we also know that the smallest eigenvalue is bounded by zero, $\lambda_{n}\geq0$.
Motivated by the identity presented in~(\ref{eqn:logdet_tr_identity}), the Taylor series expansion~\citep{Barry99, Zhang07} may be employed for evaluating the log-determinant of matrices having eigenvalues bounded between zero and one.
In particular, this approach relies on the following logarithm identity,
\begin{equation}
\centering
\log\left(I - A\right) = - \sum_{k=1}^{\infty} \frac{A^k}{k}.
\end{equation}

While the infinite summation is not explicitly computable in finite time, this may be approximated by computing a truncated series instead.
Furthermore, given that the trace of matrices is additive, we find
\begin{equation}
\centering
\Tr\bigl(\log\left(I - A\right)\bigr) \approx - \sum_{k=1}^{m} \frac{\Tr\left(A^k\right)}{k}.
\end{equation}

The $\Tr(A^k)$ term can be computed efficiently and recursively by propagating $\bigO(n^2)$ vector-matrix multiplications in a stochastic trace estimation scheme.
To compute $\Tr(\log(K))$ we simply set $A = I - K$.

There are two sources of error associated with this approach; the first due to stochastic trace estimation, and the second due to truncation of the Taylor series.
In the case of covariance matrices, the smallest eigenvalue tends to be very small, which can be verified by \cite{Weyl12} and \cite{Silverstein86}'s observations on the eigenspectra of covariance matrices.
This leads to $A^k$ decaying slowly as $k \rightarrow \infty$.

In light of the above, standard Taylor approximations to the log-determinant of covariance matrices are typically unreliable, even when the exact traces of matrix powers are available.
This can be verified analytically based on results from kernel theory, which state that the approximate rate of decay for the eigenvalues of positive definite kernels which are $\nu$-continuous is $\bigO(n^{-\nu - 0.5})$~\citep{Weyl12, Wathen15}.
Combining this result with the absolute error, $E(\lambda)$, of the truncated Taylor approximation we find
\begin{align*}
\mathbf{E}\left[E\left(\lambda\right)\right] &= \mathcal{O}\left(\int_0^1 \lambda^{\nu + 0.5}\biggl(\log\left(\lambda\right) - \sum_{j=1}^m \frac{\lambda^j}{j}\biggr) \mathrm{d}\lambda\right) \\
&= \mathcal{O}\left(\int_0^1 \lambda^{\nu + 0.5}\sum_{j=m}^\infty \frac{\lambda^j}{j} 	\mathrm{d}\lambda\right) \\
&= \mathcal{O}\left(\frac{\Psi^{(0)}\left(m + \nu + 1.5\right) - \Psi^{(0)}\left(m\right)}{\nu + 1.5}\right)\text{,}
\end{align*}

where $\Psi^{(0)}(\cdot)$ is the Digamma function.
In Figure \ref{TaylorError}, we plot the relationship between the order of the Taylor approximation and the expected absolute error.
It can be observed that irrespective of the continuity of the kernel, the error converges at a rate of $\mathcal{O}(n^{-1})$.

\section{THE PROBABILISTIC NUMERICS APPROACH}

We now propose a probabilistic numerics \citep{Hennig15} approach: we'll re-frame a numerical computation (in this case, trace estimation) as probabilistic inference. 
Probabilistic numerics usually requires distinguishing: an appropriate latent function; data and; the ultimate object of interest. 
Given the data, a posterior distribution is calculated for the object of interest. 
For instance, in numerical integration, the latent function is the integrand, $f$, the data are evaluations of the integrand, $f(x)$, and the object of interest is the value of the integral, $\int f(x) p(x) \mathrm{d}x$ (see $\S~\ref{sec:BQ}$ for more details). 
In this work, our latent function is the distribution of eigenvalues of $A$, the data are noisy observations of $\Tr(A^k)$, and the object of interest is $\log(\Det(K))$.
For this object of interest, we are able to provide both expected value and variance. 
That is, although the Taylor approximation to the log-determinant may be considered unsatisfactory, the intermediate trace terms obtained when raising the matrix to higher powers may prove to be informative if considered as observations within a probabilistic model.

\subsection{RAW MOMENT OBSERVATIONS}

We wish to model the eigenvalues of $A$ from noisy observations of $\Tr\left(A^k\right)$ obtained through stochastic trace estimation, with the ultimate goal of making predictions on the infinite series of the Taylor expansion.
Let us assume that the eigenvalues are i.i.d.\ random variables drawn from $P(\lambda_i = x)$, a probability distribution over $x \in [0, 1]$. In this setting $\Tr(A) = n \mathbf{E}_{x}[P(\lambda_i = x)]$, and more generally $\Tr\left(A^k\right) = n \mathbf{R}_{x}^{(k)}[P(\lambda_i = x)]$, where $\mathbf{R}_{x}^{(k)}$ is the $k^{th}$ raw moment over the $x$ domain.
The raw moments can thus be computed as,
\begin{equation}\label{eqn:raw_moments}
\centering
\mathbf{R}_{x}^{(k)}\left[P\left(\lambda_i = x\right)\right] = \int_0^1 x^k P\left(\lambda_i = x\right) \mathrm{d}x.
\end{equation}

Such a formulation is appealing because if $P\left(\lambda_i = x\right)$ is modeled as a Gaussian process, the required integrals may be solved analytically using Bayesian Quadrature.

\subsubsection{Bayesian Quadrature}\label{sec:BQ}

Gaussian processes~\citep[GP{s};][]{Rasmussen06} are a powerful Bayesian inference method defined over functions $X \rightarrow \mathbb{R}$, such that the distribution of functions over any finite subset of the input points $X = \{\xvect_1, \ldots, \xvect_n\}$ is a multivariate Gaussian distribution.
Under this framework, the moments of the conditional Gaussian distribution for a set of predictive points, given a set of labels $\yvect = (y_1, \ldots, y_n)^{\top}$, may be computed as
\begin{equation}\label{eqn:post_mean}
\mu = \mu_0 + K_*^\top K^{-1}(\yvect - \mu_0),
\end{equation}
\begin{equation}\label{eqn:post_var}
\Sigma = K_{*,*} - K_*^\top K^{-1}K_*,
\end{equation}

with $\mu$ and $\Sigma$ denoting the posterior mean and variance, and $K$ being the $n \times n$ covariance matrix for the observed variables $\{\xvect_i, y_i; i \in (1,2,\dots n)\}$. The latter is computed as $\kappa(\xvect,\xvect')$ for any pair of points $\xvect, \xvect' \in X$.
Meanwhile, $K_*$ and  $K_{*,*}$ respectively denote the covariance between the observable and the predictive points, and the prior over the predicted points.
Note that $\mu_0$, the prior mean, may be set to zero without loss of generality.

Bayesian Quadrature~\citep[BQ;][]{OHagan91} is primarily concerned with performing integration of potentially intractable functions.
In this work, we limit our discussion to the setting where the integrand is modeled as a GP,
\[
\int p(x)\, f(x)\, \mathrm{d}x, \quad f \sim \GP(\mu, \Sigma),
\]
where $p(x)$ is some measure with respect to which we are integrating.
A full discussion of BQ may be found in~\cite{OHagan91} and \cite{Rasmussen02}; for the sake of conciseness, we only state the result that the integrals may be computed by integrating the covariance function with respect to $p(x)$ for both $K_{*}$,
\[
\kappa\left(\int x \mathrm{d}x, x'\right) = \int p\left(x\right) \kappa\left(x, x'\right) \mathrm{d}x,
\]

and $K_{*,*}$,
\[
\kappa\left(\int x \mathrm{d}x, \int x' \mathrm{d}x'\right) = \iint p\left(x\right) \kappa\left(x, x'\right) p\left(x'\right) \mathrm{d}x \mathrm{d}x'.
\]


\subsection{KERNELS FOR RAW MOMENTS AND INFERENCE ON THE LOG-DETERMINANT}

Recalling (\ref{eqn:raw_moments}), if $P(\lambda_i = x)$ is modeled using a GP, in order to include observations of $\mathbf{R}_{x}^{(k)}[P(\lambda_i = x)]$, denoted as $\mathbf{R}_{x}^{(k)}$, we must be able to integrate the kernel with respect to the polynomial in $x$,
\begin{equation}
\centering
\kappa\left(\mathbf{R}_{x}^{(k)}, x'\right) = \int_0^1 x^k \kappa\left(x, x'\right) \mathrm{d}x,
\end{equation}
\begin{equation}
\kappa\left(\mathbf{R}_{x}^{(k)}, \mathbf{R}_{x'}^{(k')}\right) = \int_0^1  \int_0^1 x^k \kappa\left(x, x'\right) x'^{k'} \mathrm{d}x \mathrm{d}x' .
\end{equation}

Although the integrals described above are typically analytically intractable, certain kernels have an elegant analytic form which allows for efficient computation.
In this section, we derive the raw moment observations for a histogram kernel, and demonstrate how estimates of the log-determinant can be obtained.
An alternate polynomial kernel is described in Appendix~\ref{app:poly_kern}.

\subsubsection{Histogram Kernel}\label{sec:hist_kern}

The entries of the histogram kernel, also known as the piecewise constant kernel, are given by $\kappa(x,x') = \sum_{j=0}^{1-m} \mathcal{H}(\frac{j}{m}, \frac{j+1}{m}, x, x')$, where
\[
\mathcal{H}\left(\frac{j}{m}, \frac{j+1}{m}, x, x'\right) = 
\begin{cases}
1 &  x,x' \in \left[\frac{j}{m}, \frac{j+1}{m}\right]\\
0 & \text{otherwise}
\end{cases}
\text{.}
\]

Covariances between raw moments may be computed as follows:
\begin{small}
\begin{equation}
\centering
\begin{split}
\kappa\left(\mathbf{R}_{x}^{(k)}, x'\right) &= \int_0^1 x^k\kappa\left(x, x'\right) \mathrm{d}x \\
& = \frac{1}{k+1}\left(\left(\frac{j+1}{m}\right)^{k+1} - \left(\frac{j}{m}\right)^{k+1}\right)\text{,}
\end{split}
\end{equation}
\end{small}

where in the above $x$ lies in the interval $\left[ \frac{j}{m}, \frac{j+1}{m}\right]$. Extending this to the covariance function between raw moments we have,
\begin{small}
\begin{equation}
\begin{split}\label{eqn:k_obs_obs}
\kappa\left(\mathbf{R}_{x}^{(k)}, \mathbf{R}_{x'}^{(k')}\right) &= \int_0^1\int_0^1 x^k, x'^{k'} \kappa\left(x, x'\right) \mathrm{d}x \mathrm{d}x' \\
& \hspace{-15mm}= \sum_{j=0}^{m-1} \prod_{\bar{k} \in (k, k')}\frac{1}{\left(\bar{k}+1\right)}\left(\left(\frac{j+1}{m}\right)^{\bar{k}+1} - \left(\frac{j}{m}\right)^{\bar{k}+1}\right)\text{.}
\end{split}
\end{equation}
\end{small}

This simple kernel formulation between observations of the raw moments compactly allows us to perform inference over $P(\lambda_i = x)$.
However, the ultimate goal is to predict $\log(\Det(K))$, and hence $\sum_{i=1}^\infty \frac{\Tr\left(A^k\right)}{k}$.
This requires a seemingly more complex set of kernel expressions; nevertheless, by propagating the implied infinite summations into the kernel function, we can also obtain the closed form solutions for these terms,

\begin{small}
\begin{equation}\label{eqn:k_pred_obs}
\begin{split}
\kappa\left(\sum_{k=1}^\infty \frac{\mathbf{R}_{x}^{(k)}}{k},  \mathbf{R}_{x'}^{(k')}\right)  &=\sum_{j=0}^{m-1}\frac{1}{k'+1}\left(\left(\frac{j+1}{m}\right)^{k+1} - \right.\\
&\left.\left(\frac{j}{m}\right)^{k+1}\right)\left(S\left(\frac{j+1}{m}\right) - S\left(\frac{j}{m}\right)\right)
\end{split}
\end{equation}
\end{small}\\
\begin{small}
\begin{equation}\label{eqn:k_pred_pred}
\begin{split}
\kappa\left(\sum_{k=1}^\infty \frac{\mathbf{R}_{x}^{(k)}}{k},  \sum_{k'=1}^\infty \frac{\mathbf{R}_{x'}^{(k')}}{k'}\right) & = \sum_{j=0}^{m-1}\left(S\left(\frac{j+1}{m}\right) - S\left(\frac{j}{m}\right)\right)^{2}
\end{split}
\end{equation}
\end{small}

where $S(\alpha) = \sum_{k=1}^{\infty}\frac{\alpha^{k+1}}{k(k+1)}$, which has the convenient identity for $0<\alpha<1$,
\[
S(\alpha) = \alpha + (1 - \alpha)\log(1-\alpha).
\]

Following the derivations presented above, we can finally go about computing the prediction for the log-determinant, and its corresponding variance, using the GP posterior equations given in (\ref{eqn:post_mean}) and (\ref{eqn:post_var}).
This can be achieved by replacing the terms $K_*$ and $K_{*,*}$ with the constructions presented in (\ref{eqn:k_pred_obs}) and (\ref{eqn:k_pred_pred}), respectively.
The entries of $K$ are filled in using (\ref{eqn:k_obs_obs}), whereas $\yvect$ denotes the noisy observations of $\Tr\left(A^k\right)$.

\subsubsection{Prior Mean Function}\label{sec:prior_trunc}

While GPs, and in this case BQ, can be applied with a zero mean prior without loss of generality, it is often beneficial to have a mean function as an initial starting point.
If $P(\lambda_i = x)$ is composed of a constant mean function $g(\lambda_i = x)$, and a GP is used to model the residual, we have that
$$
P\left(\lambda_i = x\right) = g\left(\lambda_i = x\right) + f\left(\lambda_i = x\right).
$$

\noindent The previously derived moment observations may then be decomposed into,
\begin{equation}
\begin{split}
\int x^k P\left(\lambda_i = x\right) \mathrm{d}x &= \int x^k g\left(\lambda_i = x\right) \mathrm{d}x\\
& + \int x^kf\left(\lambda_i = x\right)\mathrm{d}x.
\end{split}
\end{equation}

Due to the domain of $P\left(\lambda_i = x\right)$ lying between zero and one, we set a Beta distribution as the prior mean, which has some convenient properties.
First, it is fully specified by the mean and variance of the distribution, which can be computed using the trace and Frobenius norm of the matrix.
Secondly, the $r$-{th} raw moment of a Beta distribution parameterized by $\alpha$ and $\beta$ is
$$
\mathbf{R}_{x}^{(k)}\left[g\left(\lambda_i = x\right)\right] = \frac{\alpha + r}{\alpha + \beta + r}\text{,}
$$
which is straightforward to compute.

In consequence, the expectation of the logarithm of random variables and, hence,  the `prior' log determinant yielded by $g\left(\lambda_i = x\right)$ can be computed as
\begin{equation}
\mathbf{E}[\log(X); X \sim g(\lambda_i = x)] = \phi(\alpha) - \phi(\alpha+\beta)\text{.}
\end{equation}

This can then simply be added to the previously derived GP expectation of the log-determinant.

\subsubsection{Using Bounds on the Log-Determinant}\label{sec:hyp_optim}

As with most GP specifications, there are hyperparameters associated with the prior and the kernel.
The optimal settings for these parameters may be obtained via optimization of the standard GP log marginal likelihood, defined as
$$
\text{LML}_{GP} = -\frac{1}{2} \yvect^\top K^{-1}\yvect - \frac{1}{2}\log(\Det(K)) + \mathrm{const}.
$$

Borrowing from the literature on bounds for the log-determinant of a matrix, as described in Appendix \ref{app:bounds}, we can also exploit such upper and lower bounds to truncate the resulting GP distribution to the relevant domain, which is expected to greatly improve the predicted log-determinant.
These additional constraints can then be propagated to the hyperparameter optimization procedure by incorporating them into the likelihood function via the product rule, as follows:
$$
\text{LML} = \text{LML}_{GP} + \log\left(\Phi\left(\frac{a - \hat{\mu}}{\hat{\sigma}}\right) - \Phi\left(\frac{b - \hat{\mu}}{\hat{\sigma}}\right)\right),
$$

\noindent with $a$ and $b$ representing the upper and lower log-determinant bounds respectively, $\hat{\mu}$ and $\hat{\sigma}$ representing the posterior mean and standard deviation, and $\Phi(\cdot)$ representing the Gaussian cumulative density function.
Priors on the hyperparameters may be accounted for in a similar way.

\subsubsection{Algorithm Complexity and Recap}

Due to its cubic complexity, GP inference is typically considered detrimental to the scalability of a model.
However, in our formulation, the GP is only being applied to the noisy observations of $\Tr\left(A^k\right)$, which rarely exceed the order of tens of points.
As a result, given that we assume this to be orders of magnitude smaller than the dimensionality $n$ of the matrix $K$, the computational complexity is dominated by the matrix-vector operations involved in stochastic trace estimation, i.e.\ $\mathcal{O}(n^2)$ for dense matrices and $\mathcal{O}(ns)$ for $s$-sparse matrices.

The steps involved in the procedure described within this section are summarized as pseudo-code in Algorithm~\ref{alg:probnum}.
The input matrix $A$ is first normalized by using Gershgorin intervals to find the largest eigenvalue (line 1), and the expected bounds on the log-determinant (line 2) are calculated using matrix theory~(Appendix~\ref{app:bounds}).
The noisy Taylor observations  up to an expansion order \textsc{m} (lines 3-4), denoted here as $\yvect$, are then obtained through stochastic trace estimation, as described in $\S~\ref{sec:taylor}$.
These can be modeled using a GP, where the entries of the kernel matrix $K$ (lines 5-7) are computed using~(\ref{eqn:k_obs_obs}).
The kernel parameters are then tuned as per $\S~\ref{sec:hyp_optim}$ (line 8).
Recall that we seek to make a prediction for the infinite Taylor expansion, and hence the exact log-determinant.
To this end, we must compute $K_*$ (lines 9-10) and $k_{*,*}$ (line 11) using (\ref{eqn:k_pred_obs}) and (\ref{eqn:k_pred_pred}), respectively.
The posterior mean and variance (line 12) may then be evaluated by filling in (\ref{eqn:post_mean}) and (\ref{eqn:post_var}).
As outlined in the previous section, the resulting posterior distribution can be truncated using the derived bounds to obtain the final estimates for the log-determinant and its uncertainty (line 13).

\algrenewcommand\algorithmicindent{1.3em}
\renewcommand{\algorithmicrequire}{\textbf{Input:}}
\renewcommand{\algorithmicensure}{\textbf{Output:}}

\begin{algorithm}
\caption{Computing log-determinant and uncertainty using probabilistic numerics}\label{alg:probnum}
\begin{algorithmic}[1]
\vspace{0.5em}
\Require PSD matrix $A\in\mathbb{R}^{n \times n}$, raw moments kernel $\kappa$, expansion order $\textsc{m}$, and random vectors \textsc{z}
\Ensure Posterior mean $\textsc{mTrn}$, and uncertainty $\textsc{vTrn}$
\vspace{0.5em}
\State $A \gets \textsc{normalize}(A)$
\State $\textsc{bounds} \gets \textsc{getBounds}(A)$
\For{$i \gets 1 \textrm{ to } \textsc{m}$}
\State $\yvect_i \gets \textsc{stochasticTaylorObs}(A, i,\textsc{z})$
\EndFor 
\For{$i \gets 1 \textrm{ to } \textsc{m}$}
\For{$j \gets 1 \textrm{ to } \textsc{m}$}
\State $K_{ij} \gets \kappa(i, j)$
\EndFor
\EndFor
\State $\kappa, K \gets \textsc{tuneKernel}(K, \yvect, \textsc{bounds})$ 
\For{$i \gets 1 \textrm{ to } \textsc{m}$}
\State $K_{*,i} \gets \kappa(*, i)$
\EndFor 
\State $k_{*,*} \gets \kappa(*, *)$
\State $\textsc{mExp}, \textsc{vExp} \gets \textsc{gpPred}(\yvect, K, K_{*}, k_{*,*})$
\State $\textsc{mTrn}, \textsc{vTrn} \gets \textsc{trunc}(\textsc{mExp}, \textsc{vExp}, \textsc{bounds})$
\end{algorithmic}
\end{algorithm}



\section{EXPERIMENTS}

\begin{figure*}[t!]
	\centering
	\includegraphics[width=1.\textwidth]{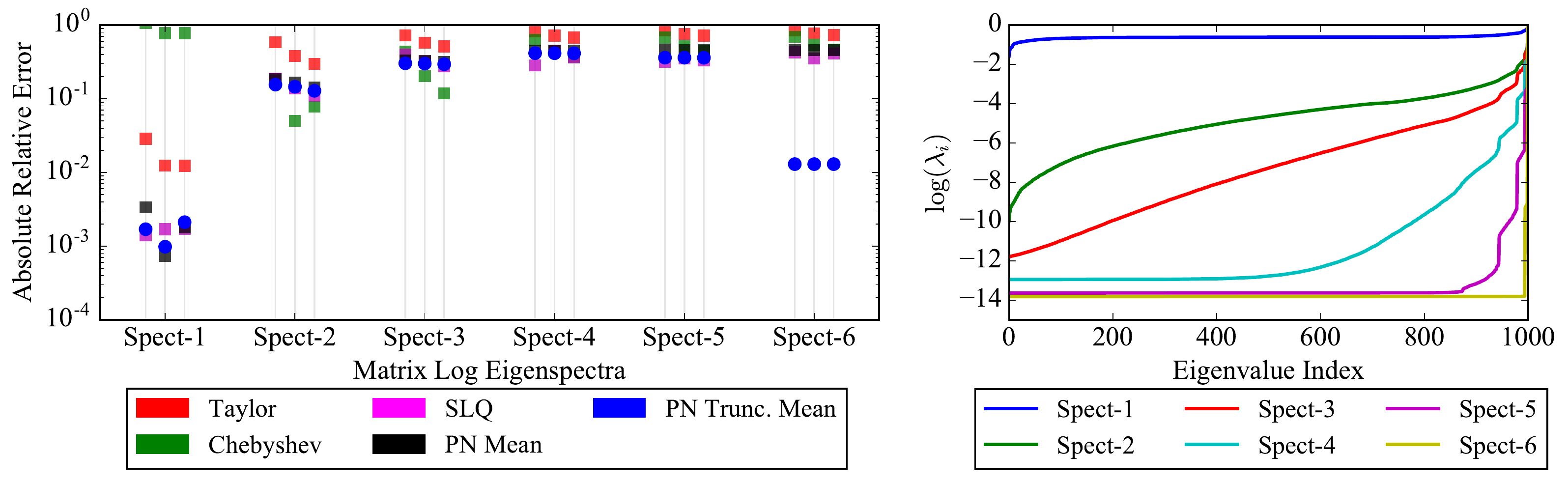}
	\caption{Empirical performance of 6 covariances described in $\S~\ref{sec:syn_exp}$.
	The right figure displays the log eigenspectrum of the matrices and their respective indices. The left figure displays the relative performance of the algorithms for the stochastic trace estimation order set to 5, 25 and 50 (from left to right respectively).
	}
	\label{fig:eigspect}
\end{figure*}

In this section, we show how the appeal of this formulation extends beyond its intrinsic novelty, whereby we also consistently obtain performance improvements over competing techniques.
We set up a variety of experiments for assessing the model performance, including both synthetically constructed and real matrices.
Given the model's probabilistic formulation, we also assess the quality of the uncertainty estimates yielded by the model.
We conclude by demonstrating how this approach may be fitted within a practical learning scenario.

We compare our approach against several other estimations to the log-determinant, namely approximations based on Taylor expansions, Chebyshev expansions and Stochastic Lanczos quadrature. The Taylor approximation has already been introduced in $\S~\ref{sec:taylor}$, and we briefly describe the others below.

\textbf{Chebyshev Expansions:} This approach utilizes the $m$-degree Chebyshev polynomial approximation to the function $\log\left(I - A\right)$ \citep{Han15, Boutsidis15, Peng15},
\begin{equation}
\centering
\text{Tr}\left(\log\left(I - A\right)\right) \approx \sum_{k=0}^{m} c_k \text{Tr}\left(T_k\left(A\right)\right),
\end{equation}
where $T_k(x) = AT_{k-1}\left(A\right) - T_{k-2}\left(A\right)$ starting with $T_0(A) = 1$ and $T_0\left(A\right) = 2*A - 1$, and $c_k$ is defined as
\begin{equation}
\begin{split}
c_k &= \frac{2}{n+1}\sum_{i=0}^n \log\left(I - x_i\right) T_k\left(x_i\right), \\
x_i &= \cos \left( \frac{\left(i+\frac{1}{2}\right)\pi}{n+1} \right).
\end{split}
\end{equation}
The Chebyshev approximation is appealing as it gives the best $m$-degree polynomial approximation of $\log\left(I - x\right)$ under the $L_\infty$-norm.
The error induced by general Chebyshev polynomial approximations has also been thoroughly investigated~\citep{Han15}.

\textbf{Stochastic Lanczos Quadrature:} This approach~\citep{Ubaru16} relies on stochastic trace estimation to approximate the trace using the identity presented in (\ref{eqn:logdet_tr_identity}).
If we consider the eigendecomposition of matrix $A$ into $Q\Lambda Q^\top$, the quadratic form in the equation becomes
\begin{align*}
\centering
\begin{split}
{\mathbf{r}^{(i)}}^\top\log(A){\mathbf{r}^{(i)}} &= {\mathbf{r}^{(i)}}^\top Q\log\left(\Lambda\right)Q^\top{\mathbf{r}^{(i)}} \\
&= \sum_{k=1}^{n}\log\left(\lambda_k\right)\mu_k^2
\end{split}\textsc{,}
\end{align*}
where $\mu_k$ denotes the individual components of $Q^\top{\mathbf{r}^{(i)}}$.
By transforming this term into a Riemann-Stieltjes integral $\int_a^b\log(t)\mathrm{d}\mu(t)$, where $\mu(t)$ is a piecewise constant function~\citep{Ubaru16}, we can approximate it as
$$
\int_a^b\log(t)\mathrm{d}\mu(t) \approx \sum_{j=0}^m\omega_j\log\left(\theta_j\right)\textsc{,}
$$
where $m$ is the degree of the approximation, while the sets of $\omega$ and $\theta$ are the parameters to be inferred using Gauss quadrature.
It turns out that these parameters may be computed analytically using the eigendecomposition of the low-rank tridiagonal transformation of $A$ obtained using the Lanczos algorithm~\citep{Paige72}.
Denoting the resulting eigenvalues and eigenvectors by $\theta$ and $y$ respectively, the quadratic form may finally be evaluated as,
\begin{equation}
{\mathbf{r}^{(i)}}^\top\log\left(A\right){\mathbf{r}^{(i)}} \approx \sum_{j=0}^m\tau_j^2\log\left(\theta_j\right)\text{,}
\end{equation}
with $\tau_j=\left[e_1^Ty_j\right]$.

\subsection{SYNTHETICALLY CONSTRUCTED MATRICES}\label{sec:syn_exp}

Previous work on estimating log-determinants have implied that the performance of any given method is closely tied to the shape of the eigenspectrum for the matrix under review.
As such, we set up an experiment for assessing the performance of each technique when applied to synthetically constructed matrices whose eigenvalues decay at different rates.
Given that the computational complexity of each method is dominated by the number of matrix-vector products (\mvp{s}) incurred, we also illustrate the progression of each technique for an increasing allowance of \mvp{s}. All matrices are constructed using a Gaussian kernel evaluated over 1000 input points.

As illustrated in Figure~\ref{fig:eigspect}, the estimates returned by our approach are consistently on par with (and frequently superior to) those obtained using other methods.
For matrices having slowly-decaying eigenvalues, standard Chebyshev and Taylor approximations fare quite poorly, whereas SLQ and our approach both yield comparable results.
The results become more homogeneous across methods for faster-decaying eigenspectra, but our method is frequently among the top two performers.
For our approach, it is also worth noting that truncating the GP using known bounds on the log-determinant indeed results in superior posterior estimates.
This is particularly evident when the eigenvalues decay very rapidly.
Somewhat surprisingly, the performance does not seem to be greatly affected by the number of budgeted \mvp{s}.

\subsection{UFL SPARSE DATASETS}

\begin{figure}
	\centering
	\includegraphics[width=.4\textwidth]{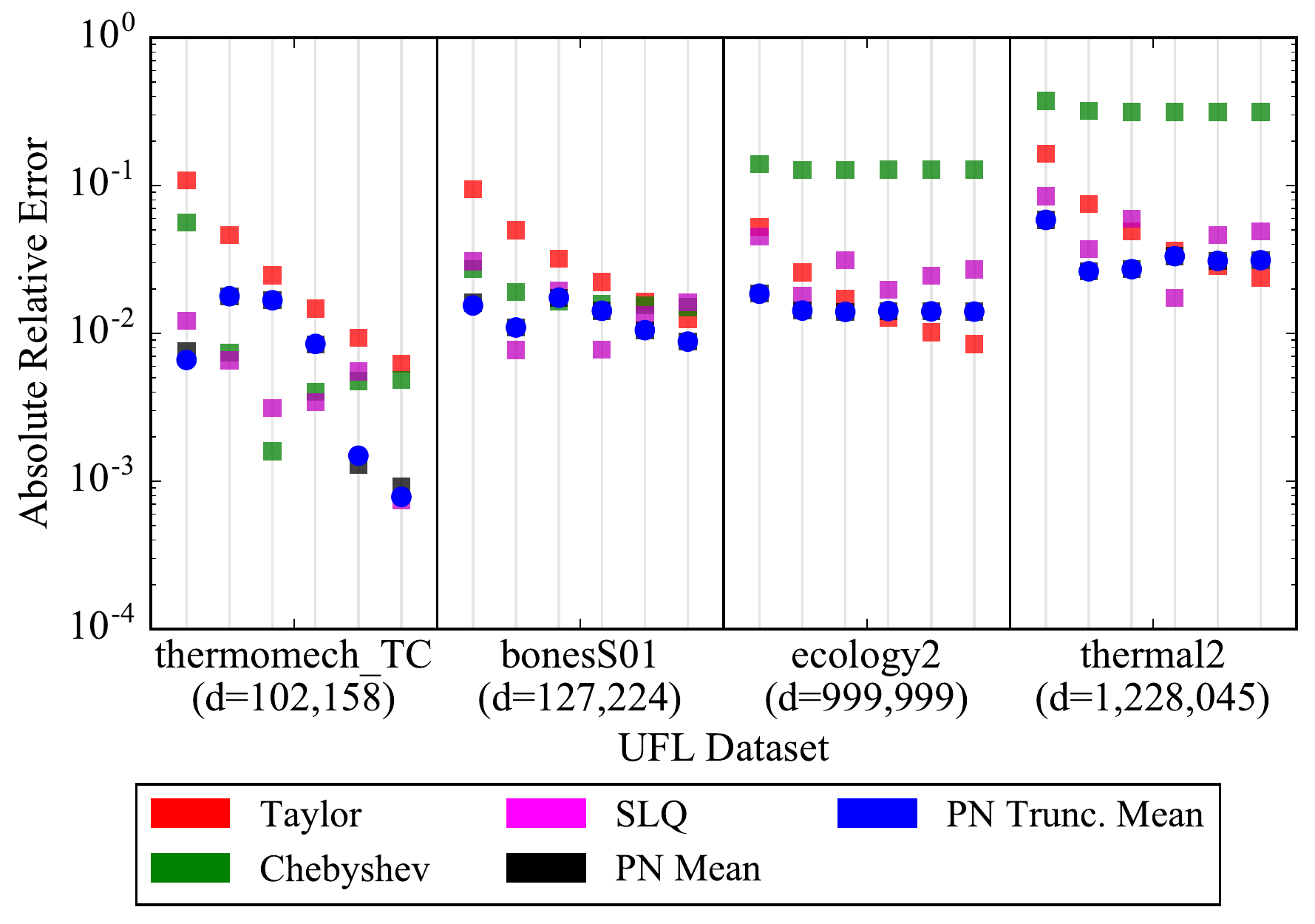}
	\caption{Methods compared on a variety on UFL Sparse Datasets.
	Each dataset was ran the matrix approximately raised to the power of 5, 10, 15, 20, 25 and 30 (left to right) using stochastic trace estimation.
	}
	\label{fig:ufl}
\end{figure}

Although we have so far limited our discussion to covariance matrices, our proposed method is amenable to any positive semi-definite matrix.
To this end, we extend the previous experimental set-up to a selection of real, sparse matrices obtained from the SuiteSparse Matrix Collection~\citep{Davis11}.
Following~\cite{Ubaru16}, we list the true values of the log-determinant reported in~\cite{Boutsidis15}, and compare all other approaches to this baseline. 

The results for this experiment are shown in Figure~\ref{fig:ufl}.
Once again, the estimates obtained using our probabilistic approach achieve comparable accuracy to the competing techniques, and several improvements are noted for larger allowances of \mvp{s}.
As expected, the SLQ approach generally performs better than Taylor and Chebyshev approximations, especially for smaller computational budgets.
Even so, our proposed technique consistently appears to have an edge across all datasets.

\subsection{UNCERTAINTY QUANTIFICATION}

\begin{figure}[h!]
\begin{center}
	\includegraphics[width=.4\textwidth]{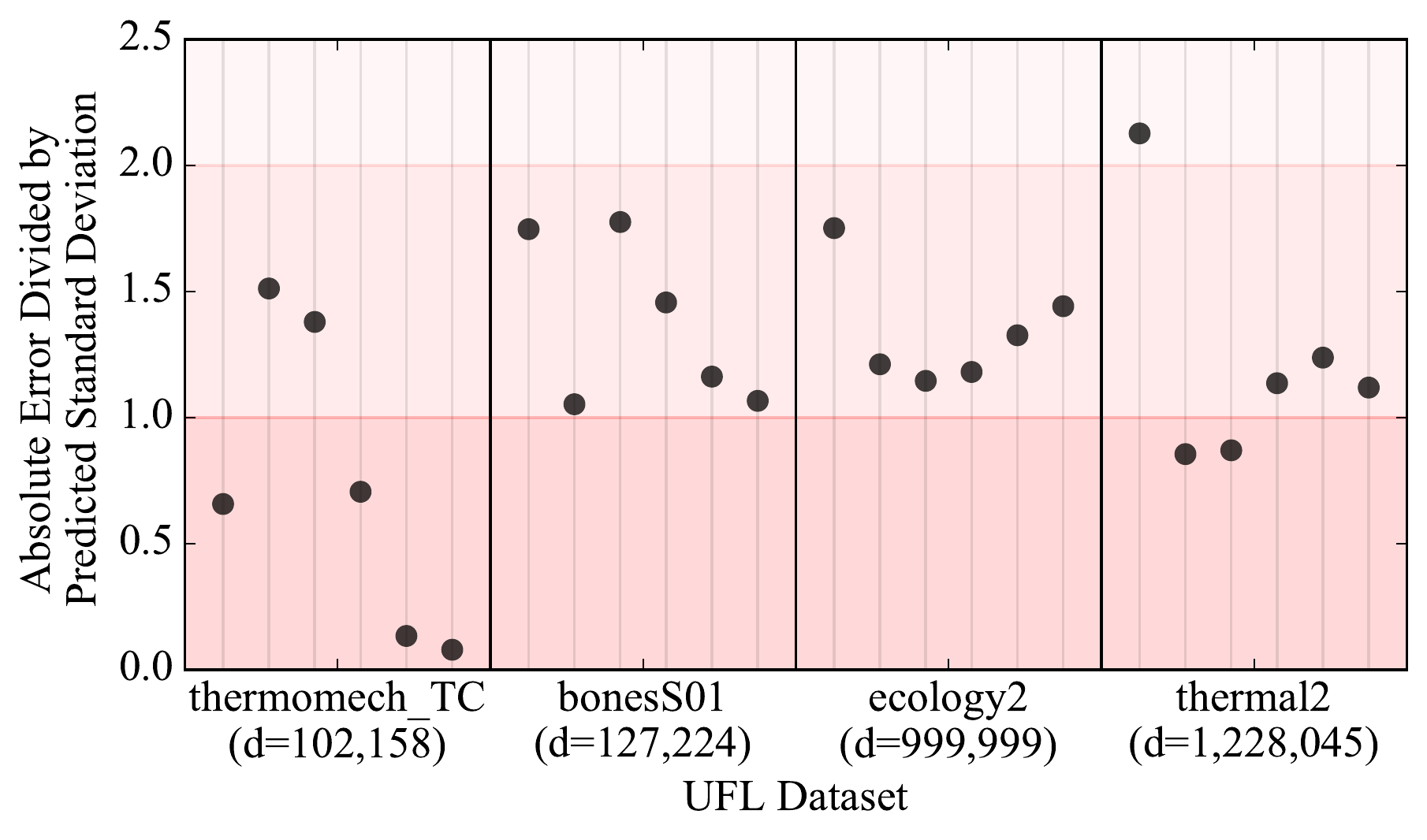}
	\caption{Quality of uncertainty estimates on UFL datasets, measured as the ratio of the absolute error to the predicted variance.
	As before, results are shown for increasing computational budgets (\mvp{s}).
	The true value lay outside 2 standard deviations in only one of 24 trials.}
	\label{fig:ufl_unc}
\end{center}
\end{figure}

One of the notable features of our proposal is the ability to quantify the uncertainty of the predicted log-determinant, which can be interpreted as an indicator of the quality of the approximation.
Given that none of the other techniques offer such insights to compare against, we assess the quality of the model's uncertainty estimates by measuring the ratio of the absolute error to the predicted standard deviation (uncertainty).
For the latter to be meaningful, the error should ideally lie within only a few multiples of the standard deviation.

In Figure~\ref{fig:ufl_unc}, we report this metric for our approach when using the histogram kernel.
We carry out this evaluation over the matrices introduced in the previous experiment, once again showing how the performance varies for different \mvp allowances.
In all cases, the absolute error of the predicted log-determinant is consistently bounded by at most twice the predicted standard deviation, which is very sensible for such a probabilistic model.

\subsection{MOTIVATING EXAMPLE}

Determinantal point processes~\citep[DPP{s};][]{Macchi75} are stochastic point processes defined over subsets of data such that an established degree of repulsion is maintained.
A DPP, $\mathcal{P}$, over a discrete space $y \in \{1,\dots, n\}$ is a probability measure over all subsets of $y$ such that
\[
\mathcal{P}(A \in y) = \Det(K_A),
\]

where $K$ is a positive definite matrix having all eigenvalues less than or equal to 1.
A popular method for modeling data via $K$ is the \emph{L-ensemble} approach \citep{borodin2009determinantal}, which transforms kernel matrices, $L$, into an appropriate $K$,
\[
K = (L+I)^{-1}L.
\]

The goal of inference is to correctly parameterize $L$ given observed subsets of $y$, such that the probability of unseen subsets can be accurately inferred in the future.

\begin{figure}[t!]
	\centering
	\includegraphics[width=.4\textwidth]{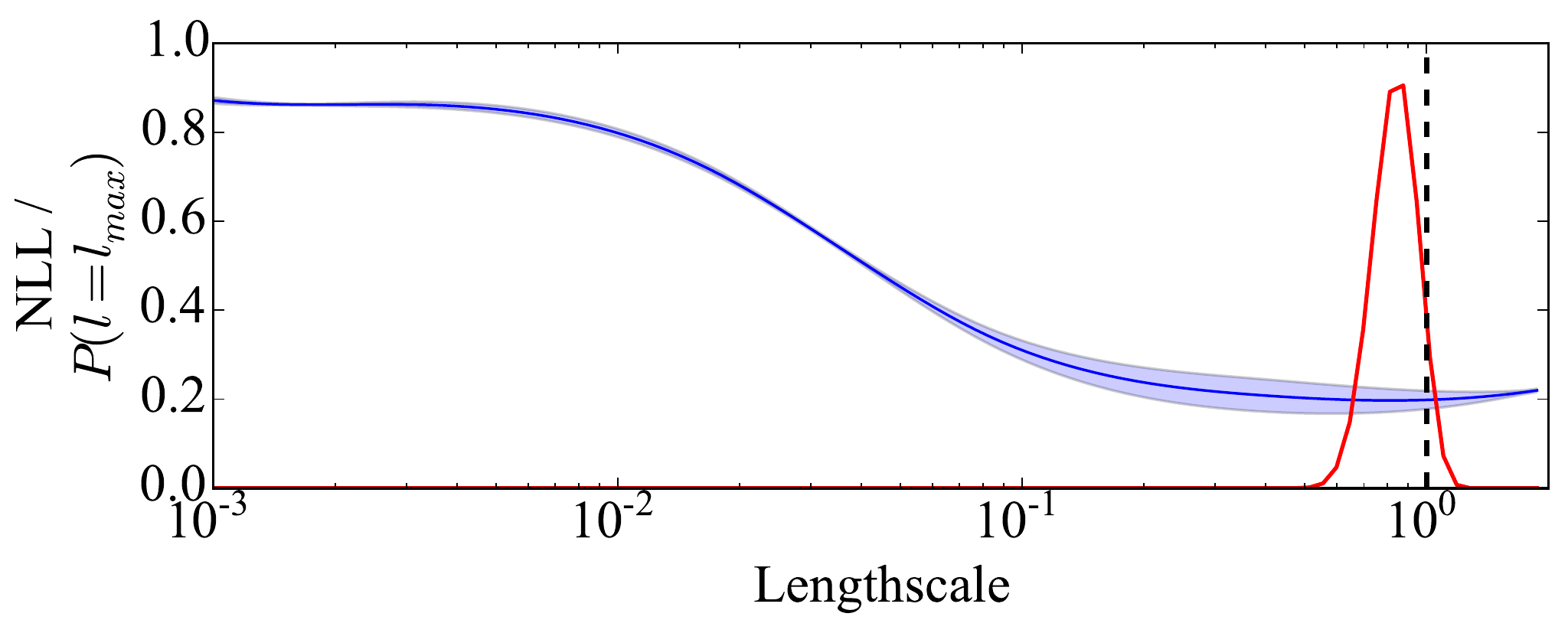}
	\caption{The rescaled Negative log likelihood (NLL) of DPP with varying length scale (blue) and probability of maximum likelihood (red). Cubic interpolation was used between inferred likelihood observations. Ten samples, $z$, were taken to polynomial order 30.
	\vspace{-2mm}
	}
	\label{fig:dpp}
\end{figure}

Given that the log-likelihood term of a DPP requires the log-determinant of $L$, na\"{i}ve computations of this term are intractable for large sample sizes.
In this experiment, we demonstrate how our proposed approach can be employed to the purpose of parameter optimization for large-scale DPPs.
In particular, we sample points from a DPP defined on a lattice over $[-1,1]^5$, with one million points at uniform intervals.
A Gaussian kernel with lengthscale parameter $l$ is placed over these points, creating the true $L$.
Subsets of the lattice points can be drawn by taking advantage of the tensor structure of $L$, and we draw five sets of 12,500 samples each.
For a given selection of lengthscale options, the goal of this experiment is to confirm that the DPP likelihood of the obtained samples is indeed maximized when $L$ is parameterized by the true lengthscale, $l$. 
As shown in Figure \ref{fig:dpp}, the computed uncertainty allows us to derive a distribution over the true lengthscale which, despite using few matrix-vector multiplications, is very close to the optimal.


\section{CONCLUSION}

In a departure from conventional approaches for estimating the log-determinant of a matrix, we propose a novel probabilistic framework which provides a Bayesian perspective on the literature of matrix theory and stochastic trace estimation.
In particular, our approach enables the log-determinant to be inferred from noisy observations of $\Tr\left(A^k\right)$ obtained from stochastic trace estimation.
By modeling these observations using a GP, a posterior estimate for the log-determinant may then be computed using Bayesian Quadrature.
Our experiments confirm that the results obtained using this model are highly comparable to competing methods, with the additional benefit of measuring uncertainty.

We forecast that the foundations laid out in this work can be extended in various directions, such as exploring more kernels on the raw moments which permit tractable Bayesian Quadrature.
The uncertainty quantified in this work is also a step closer towards fully characterizing the uncertainty associated with approximating large-scale kernel-based models.



\subsubsection*{Acknowledgements}

Part of this work was supported by the Royal Academy of Engineering and the Oxford-Man Institute. MF gratefully acknowledges support from the AXA Research Fund. The authors would like to thank Jonathan Downing for his supportive and insightful conversation on this work.

\subsubsection*{References}


\begingroup
\renewcommand{\section}[2]{}%
\bibliographystyle{icml2017}
\setlength{\bibhang}{0pt}
\bibliography{references}
\endgroup

\appendix 
\onecolumn

\section{POLYNOMIAL KERNEL}\label{app:poly_kern}

Similar to the derivation of the histogram kernel, we can also derive the polynomial kernel for moment observations. The entries of the polynomial kernel, given by $k(x,x') = (xx' + c)^d$, can be integrated over as,

\vspace{-5mm}

\begin{center}
\begin{equation}
\begin{split}
\kappa\left(\mathbf{R}_{x}^{(k)}, x'\right) &= \int_0^1 \sum_{i=1}^d {{d}\choose{i}} x^{k+i}x'^i c^{d-i}  dx,\\
&=  \sum_{i=1}^d {{d}\choose{i}} \frac{x'^i c^{d-i} }{k+i+1}.
\end{split}
\end{equation}
\end{center}
\begin{center}
\begin{equation}
\begin{split}\kappa\left(\mathbf{R}_{x}^{(k)},  \mathbf{R}_{x'}^{(k')}\right) &= \int_0^1 \int_0^1 \sum_{i=1}^d {{d}\choose{i}} x^{k+i}x'^{k'+i} c^{d-i}  dx dx'\\
&=  \sum_{i=1}^d {{d}\choose{i}} \frac{c^{d-i} }{\left(k+i+1\right)\left(k' + i +1\right)}.
\end{split}
\end{equation}
\end{center}

\vspace{-5mm}
As with the histogram kernel, the infinite sum of the Taylor expansion can also be combined into the Gaussian process,

\vspace{-5mm}

\begin{center}
\begin{equation}
\begin{split}
\kappa\left(\sum_{k=1}^\infty \frac{\mathbf{R}_{x}^{(k)}}{k},  \mathbf{R}_{x'}^{(k')}\right) &= \frac{1}{k} \sum_{k=1}^\infty  \sum_{i=1}^d {{d}\choose{i}} \frac{c^{d-i} }{\left(k+i+1\right)\left(k' + i +1\right)}\\
&=  \sum_{i=1}^d {{d}\choose{i}} \frac{c^{d-i} \left(\Psi^{(0)}\left(i+2\right) + \gamma\right)}{\left(i+1\right)\left(k' + i +1\right)}\text{,}
\end{split}
\end{equation}
\end{center}

\begin{center}
\begin{equation}
\begin{split}
\kappa\left(\sum_{k=1}^\infty \frac{\mathbf{R}_{x}^{(k)}}{k},  \sum_{k'=1}^\infty \frac{\mathbf{R}_{x'}^{(k')}}{k'}\right) &= \frac{1}{kk'} \sum_{k=1}^\infty \sum_{k'=1}^\infty  \sum_{i=1}^d {{d}\choose{i}} \frac{c^{d-i} }{\left(k+i+1\right)\left(k' + i +1\right)}\\
&=  \sum_{i=1}^d {{d}\choose{i}} \frac{c^{d-i} \left(\Psi^{(0)}\left(i+2\right) + \gamma\right)^2}{\left(i+1\right)^2}\text{.}
\end{split}
\end{equation}
\end{center}

In the above, $\Psi^{(0)}(\cdot)$ is the Digamma function and $\gamma$ is the Euler-Mascheroni constant. We strongly believe that the polynomial and histogram kernels are not the only kernels which can analytically derived to include moment observations but act as a reasonable initial choice for practitioners.

\section{BOUNDS ON LOG DETERMINANTS}\label{app:bounds}

For the sake of completeness, we restate the bounds on the log determinants used throughout this paper \citep{Bai97}.

\begin{theorem}
	Let $A$ be an n-by-n symmetric positive definite matrix, $\mu_1 = \text{Tr}(A)$, $\mu_2 = \|A\|_F^2$ and $\lambda_i(A) \in [\alpha;\beta]$ with $\alpha > 0$, then

\vspace{-5mm}

\[\left[ \begin{array}{c}
\log\alpha \\ 
\log t \end{array} \right]^T
\left[ \begin{array}{cc}
\alpha & t  \\
\alpha^2 & t^2  \end{array} \right]
\left[ \begin{array}{c}
\mu_1 \\ 
\mu_2 \end{array} \right] \leq \text{Tr}(\log(A)) \leq
\left[ \begin{array}{c}
\log\beta \\ 
\log \bar{t} \end{array} \right]^T
\left[ \begin{array}{cc}
\beta & \bar{t}  \\
\beta^2 & \bar{t}^2  \end{array} \right]
\left[ \begin{array}{c}
\mu_1 \\ 
\mu_2 \end{array} \right]\]

where,
\vspace{-5mm}

\[t = \frac{\alpha \mu_1 - \mu_2}{\alpha n - \mu_2}, \quad \bar{t} = \frac{\beta \mu_1 - \mu_2}{\beta n - \mu_2}\]
\end{theorem}

This bound can be easily computed during the loading of the matrix as both the trace and Frobenius norm can be readily calculated using summary statistics. However, bounds on the maximum and minimum must also be derived. We chose to use Gershgorin intervals to bound the eigenvalues~\citep{Gershgorin31}.

\end{document}